\documentclass[conference]{IEEEtran}
\usepackage{cite}
\usepackage{amsmath,amssymb,amsfonts}
\usepackage{algorithmic}
\usepackage{graphicx}
\usepackage{textcomp}
\usepackage{xcolor}
\usepackage[symbol]{footmisc}
\usepackage{hyperref}
\def\BibTeX{{\rm B\kern-.05em{\sc i\kern-.025em b}\kern-.08em
    T\kern-.1667em\lower.7ex\hbox{E}\kern-.125emX}}
\begin{document}

\title{Parallelized Interactive Machine Learning on Autonomous Vehicles}

\author{
\IEEEauthorblockN{Xi Chen$^{\dagger}$}
\IEEEauthorblockA{\textit{Cell and Molecular Biology} \\
\textit{University of Kentucky}\\
Lexington, KY USA \\
billchenxi@uky.edu}
\and
\IEEEauthorblockN{Caylin Hickey$^{\dagger}$}
\IEEEauthorblockA{\textit{Computer Science} \\
\textit{University of Kentucky}\\
Lexington, KY USA \\
caylin.hickey@uky.edu}
}
\maketitle

\footnotetext[2]{These authors contributed equally to this work}

\begin{abstract}
Deep reinforcement learning (deep RL) has achieved superior performance in complex sequential tasks by learning directly from image input. A deep neural network is used as a function approximator and requires no specific state information. However, one drawback of using only images as input is that this approach requires a prohibitively large amount of training time and data for the model to learn the state feature representation and approach reasonable performance. This is not feasible in real-world applications, especially when the data are expansive and training phase could introduce disasters that affect human safety. In this work, we use a human demonstration approach to speed up training for learning features and use the resulting pre-trained model to replace the neural network in the deep RL Deep Q-Network (DQN), followed by human interaction to further refine the model. We empirically evaluate our approach by using only a human demonstration model and modified DQN with human demonstration model included in the Microsoft AirSim car simulator. Our results show that (1) pre-training with human demonstration in a supervised learning approach is better and much faster at discovering features than DQN alone; (2) initializing the DQN with a pre-trained model provides a significant improvement in training time and performance even with limited human demonstration; and (3) providing the ability for humans to supply suggestions during DQN training can speed up the network?s convergence on an optimal policy, as well as allow it to learn more complex policies that are harder to discover by random exploration.
\end{abstract}

\begin{IEEEkeywords}
deep reinforcement learning, interactive machine learning, autonomous vehicles, simulation
\end{IEEEkeywords}

\section{Introduction}
%As self-driving vehicles gain popularity and become a more viable transportation solution for their low accidents rates, it would not be surprising to see tens to hundreds of thousands of these kinds of vehicles on the road in the next 5 years. This huge market has attracted many companies to invest in the technologies involved in self-driving, such as deep learning, computer vision, data processing, and so on. However, these new technologies face many challenges in the form of road hazards, changing conditions, etc. Therefore, it will be crucial to develop methods that allow technically unskilled users to teach the algorithm in such a way as to customize the driving experience to their needs. Human-demonstration has long been the standard training approach for self-driving industry. One fairly new method is using Convolution Neural Networks (CNNs) as function approximator in the deep reinforcement learning \cite{Nvidia}. CNNs have revolutionized pattern recognition and are especially powerful in image recognition task. By using the convolution kernels to scan road images at time points of driving, it allowed few parameters need to learn compared to the total number of operations.
As self-driving vehicles gain popularity and become a more viable transportation solution for their low accident rates, it would not be surprising to see tens to hundreds of thousands of these vehicles on the road in the next 5 years. This huge potential market has attracted many companies to invest in the technologies involved in self-driving, such as deep learning, computer vision, data processing, and so on. However, these new technologies face many challenges, in the form of road hazards, changing conditions, etc. Therefore, it will be crucial to develop methods that allow technically unskilled users to teach the algorithm in a way that allows them to customize the driving experience to their needs.

Human demonstration has long been the standard training approach for the self-driving industry. One fairly new method is the use of Convolution Neural Networks (CNNs) as a function approximator in deep RL \cite{Nvidia}. CNNs have revolutionized pattern recognition and are especially powerful in image recognition tasks. Because this approach uses convolution kernels to scan road images at different driving time points, fewer parameters will need to be trained compared to the total number of operations.
 
%Deep reinforcement learning algorithms, such as DQN, suffer from poor initial performance comparing with the classic reinforcement learning algorithm since they learn \textit{tabula rasa} \cite{Sutton_1998}. This also contributes to increased training time because the network needs to learn the unspecified features besides the policy in contrast to using hand-engineered features. In addition, complex domains, like self-driving, demand a low error margin in order to avoid safety issues. These problems are non-trivial and consequential in real-world applications. 

Deep RL algorithms, such as Deep Q-Network (DQN), suffer from poor initial performance compared with the classic RL algorithm, since they start as a \textit{tabula rasa} \cite{Sutton_1998}. This also contributes to increased training time, because these algorithms need to learn the unspecified features in addition to the policy, in contrast to using handengineered features. In addition, complex domains, like autonomous driving, demand a low error margin in order to avoid safety issues. These problem are non-trivial and consequential in real-world applications.
    
%In order to use deep reinforcement learning to solve real-world problems with low error rates, there's a need for speed and accuracy. One method is by using humans to provide demonstrations. Human demonstrations have been used in reinforcement learning for a long time. \cite{Argal_2009} However, only recently has this area regain attraction as a possible way of speeding training in deep reinforcement learning. \cite{Kurin_2017} One of our contributions is to apply human driving demonstration to a DQN algorithm by providing a pre-trained CNN model with human interaction fine-tuning down the line.

In order to use deep RL to solve real-world problems with low error rates, there is a need to increase its speed and accuracy. One method is by using humans to provide demonstrations. Human demonstrations have been used in RL for a long time \cite{Argal_2009}; however, this area has only recently garnered interest as a method that may speed training in deep RL \cite{Kurin_2017}.
     
%Using an interactive machine learning method will help the self-driving vehicles to gain expertise by fine-tuning the pre-trained deep neural network that allows self-driving agents to drive and thus gain experience in an unfamiliar region without learning from scratch. Also, facing an ever-changing entity of the city/region layout, using interactive learning could avoid risks brought by unfamiliar road conditions with novel layouts since any interaction will be able to easily guide the self-driving agent at an early age by steering the wheel and taking back the driver seat.  

One contribution of this work is its illustration of the results of applying human driving demonstration to a DQN algorithm by providing a pre-trained CNN model with later fine-tuning through human interaction. Using an interactive machine learning method will help individual self-driving vehicles to gain expertise by fine-tuning the pre-trained deep neural network that allows a self-driving agent to gain experience in a unfamiliar region without learning from scratch. Interactive learning could also help avoid risks arising from unfamiliar road conditions and new layouts, since interaction will be able easily guide the self-driving agent at an early stage by steering and take back the driver seat.
    
%Our contributions by including human driving demonstration are targeting three problems: (1) feature learning via human demonstration, (2) policy learning through DQN, and (3) interactive learning for novel environments. In this work, we address the first two problems, i.e. feature and policy learning, by speeding up the pre-trained CNN model with a demonstration of human driving to learn the underlying features in the hidden layers of the network \cite{Erhan_2009,Erhan_2010,Yosinski_2014}. To address the third problem, we augment the DQN learning process by providing the ability for a teacher to provide human suggest during the episodes used by the DQN to gather training data.

By including a human driver for demonstration we target three problems: (1) feature learning via human demonstration; (2) policy learning through DQN; and (3) interactive learning for novel environments. In this work, we address the first two problems, i.e. feature and policy learning, by speeding up the pre-trained CNN model with human demonstration to learn the underlying features in the hidden layers of the network \cite{Erhan_2009,Erhan_2010,Yosinski_2014}. We address the third problem by augmenting the DQN learning process to allow a human teacher to provide suggestions during the episodes used by the DQN to gather training data.

%In addition, we structured our learning environment from simulation, which could easily mimic any real road condition and city layout using the Unreal Engine software. In addition, we tested augmenting the training process through interactive learning using the same environment. Simulated agents traditionally don'��t represent the same environmental accuracy as real vehicles, but the new simulation, which we plan to use, have brought in this consideration and provided manual tuning in the simulation, such as rigid body, forces, and torques. \cite{shah2018airsim}

The learning environment was structured as a simulationthat can easily mimic any combination of real road conditionsand city layout using the Unreal Engine software. Inaddition, we tested augmenting the training process throughinteractive learning using the same environment; simulatedagents do not traditionally represent the same environmentalaccuracy as real vehicles, but the new simulation we usedincluded manual tuning in the simulation, e.g. rigid body,forces and torques. \cite{shah2018airsim}
    
%We test our approach in both Deep Q-Network (DQN) with and without human demonstration and evaluated its performance using the AirSim car simulator in the pre-compiled neighborhood environment domain. Our results show speedups in the learning process. The improvement in the self-driving performance is really large. The generality of this approach suggests that it is feasible and necessary for deep reinforcement learning algorithms to incorporate human demonstration and interaction.

We tested our approach in both the Deep Q-Network(DQN) with and without human demonstration and evaluatedits performance using the AirSim car simulator in aneighborhood environment domain. Our results show an increasein the speed of the learning process with a large improvementin self-driving performance. The generality ofthis approach suggests that it is feasible and necessary fordeep RL algorithms to incorporate human demonstrationand interaction.

\section{Related Work}
%Although our work is not directly under transfer learning, it is similar to the transfer learning methods in deep learning. At the domain of deep neural networks for image classification, Yosinski et al. have shown the benefits speed up of learning features from existing models when the datasets are similar. \cite{Yosinski_2014} In this work, we structured a human demonstration convolution network \cite{Nvidia}, then using the pre-trained model as a source, and this CNN model is used to initialize the reinforcement learning agent's network.

Although our work does not fall directly under the umbrellaof transfer learning, it is similar to the transfer learningmethods in deep learning. At the domain of deep neuralnetworks for image classification, Yosinski et al. haveshown the benefits speed up of learning features from existingmodels when the datasets are similar \cite{Yosinski_2014}. In this work, we structured a human demonstrationconvolution network \cite{Nvidia}, then used thepre-trained model as source, and the CNN model was thenused to initialize the RL agent's network.

%Existing research work on pre-training in reinforcement learning \cite{Abtahi_2011,Anderson_2015} has shown the improvement when using a pre-trained model from similar datasets. The capability of these studies was limited by the small number of parameters learned and by the state input. In our work, we use the raw images of simulated driving as network input from the human demonstration driving. One thing worths to notice is the pre-training model needs to learn the features of states and also learning policy.

Existing research on pre-training in RL \cite{Abtahi_2011,Anderson_2015} has shown improvementwhen using a pre-trained model on similar datasets. The capabilityof these studies were limited by the small numberof parameters learned and by the state input. In our work,we used the raw images of simulation driving domain asnetwork input from the human driving demonstration. It isworth nothing that the pre-training model needs to learn thefeatures of states as well as policy.
    
%Our approach of using supervised learning for pre-training is similar to Bojarski, et al. \cite{Nvidia}. In their model, they pre-train by learning to predict the action basing on input images and minimize the loss between predict and actual action provided by the human volunteers. We use a similar approach with image frames from human demonstration as input data and the labels are provided by the action taken corresponding to each image frame. Another approach to pre-training is to learn the latent feature by using unsupervised learning through Deep Belief Network\cite{Abtahi_2011}. Although the approach is different the fundamental goal is the same: to improve learning by using pre-trained networks instead of random initialization.

Our approach of using supervised learning for pretrainingis similar to that of \cite{Nvidia}. In their model,pre-training involves learning to predict an action based oninput image and minimizing the loss between predicted andactual actions provided by human volunteers. We used asimilar approach, with image frames from human demonstratorsas input data and labels provided by the action takencorresponding to each image frame. Another approach topre-training is to learn the latent feature by using unsupervisedlearning through deep belief networks \cite{Abtahi_2011}. Although the approach is different, the fundamentalgoal is the same: to improve learning by using pretrainednetworks instead of random initialization.
    
%Other recent works leveraging human input in deep reinforcement learning include using human feedback to learn a reward function \cite{Christiano_2017} and, similar to ours, pre-trains the network with human demonstration in DQN. \cite{Hester_2017} However, these works of pre-training (combines the large margin supervised loss and the temporal difference loss) are focused on closely imitating the demonstrator. In our work, we only use the cross-entropy loss and focus on the features learning.
Other recent work leveraging human input in deep RL includethe use of human feedback to learn a reward function\cite{Christiano_2017} and, similar to our system,pre-training of a network with human demonstrationin DQN \cite{Hester_2017}. However, these examples of pretraining(combining large-margin supervised loss and temporaldifference loss) are focused on close imitation of thedemonstrator. In our work, we use only the cross-entropyloss and focus on learning features.
    
%Another work on using supervised learning for human demonstration and use learned network to initializing reinforcement learning's policy network is from Silver et al. \cite{Silver_2016}. However, their study focuses on a single domain and used a huge amount of data provided from human experts to train the supervised network To the contrast, our work will use much less training data and illustrate the usability and feasibility of such approach impacts deep reinforcement learning algorithm and how well such an approach could benefit from human demonstration. Our work shows that only small amount of data gained from a non-expert is enough for a supervised neural network to learn important feature representation for road driving from demo image frames and deep reinforcement learning such as DQN can benefit largely from the pre-trained model as initiation.
Another study, \cite{Silver_2016}, also used supervised learningfor human demonstration and learned networks to initializethe policy network for RL. However, that study focusedon a single domain and used a huge amount of data providedby human experts to train the supervised network. In contrast,our approach used a much smaller training dataset andillustrates the usability and feasibility of such an approach toaffect the deep RL algorithm. Our study shows that a smallamount of data gained from a non-expert is enough for a supervisedneural network to learn important feature representationfor driving from demo image frames; deep RL algorithmssuch as DQN can benefit from having the pre-trainedmodel as a starting point.

\section{Deep Reinforcement Learning}
%Reinforcement learning problems are normally modeled as a Markov Decision Process, represented by a tuple of values $(S, A, P, R, \gamma)$. The essence of reinforcement learning is let the agent explore an unknown environment by taking an action $a \in A$. After taking each action, agent lands to a certain state $s \in S$. A reward $r \in R(s,a)$ is given based on the action taken and the next state $s'$ agent lands in. The aim of the reinforcement learning algorithm is to let the agent learn to maximize the expected reward $R_{t} = \sum_{k=0}^{\inf} \gamma^{k}r_{t+k}$ for each state at time $t$. The importance of future and immediate rewards is determined by the discount factor $\gamma \in (0,1]$, in which 1 suggests we treat future reward as important as immediate, close to 0 vice verse.
Reinforcement learning (RL) problems are normally modeledas a Markov Decision Process, represented by a tupleof values $(S, A, P, R, \gamma)$. The essence of RL is to let theagent explore an unknown environment by taking an action$a \in A$. After taking each action, the agent lands at a certainstate $s \in S$. A reward, $r \in R(s,a)$, is given based on theaction taken and the next state, $s'$, of the agent. The aim ofthe RL algorithm is to let the agent learn to maximize theexpected reward, $R_{t} = \sum_{k=0}^{\inf} \gamma^{k}r_{t+k}$, for each state at time$t$. The importance of future and immediate rewards is determinedby the discount factor, $\gamma \in (0,1]$; a value close to 1suggests the agent should treat a future reward as important,and vice versa for values close to 0.

\subsection{Deep Q-Network}
%The Deep Q-Network algorithm is the rising star of deep reinforcement learning domain, thanks to its ability to generalize and its flexibility on solving problems in different domains. The first deep Q-network (DQN) \cite{mnih2015human} was capable to learn to solve 49 Atari games directly from the screen pixels by combining Q-learning \cite{Watkins_1992} with a deep convolution neural network. In a classic Q-learning algorithm, it learns the value of state-action pairs instead of the value of states. 
The Deep Q-Network (DQN) algorithm is the rising star ofthe deep RL domain thanks to its ability to generalize andits flexibility in solving problems in different domains. Thefirst implementation of DQN \cite{mnih2015human} was capableof learning to solve 49 Atari games directly from the screenpixels by combining Q-learning \cite{Watkins_1992}with a deep convolution neural network.

A classic Q-learning algorithm learns the value of stateactionpairs instead of the value of states:
\begin{equation} \label{eq:qlearning}
Q*(s,a)=E_{s'}[r + \gamma max_{a'}Q*(s',a')|s,a]
\end{equation}
and uses the expected discounted reward from performing actions $a$ in state $s$. The optimal policy $\pi*$ was later calculated by maximizing the Q value $Q*(s,a)=max_{\pi}Q^{\pi}(s,a)$. 

%When in the domain that state space is fairly large or continuous (e.g., in Atari games and driving), it's not feasible to compute the Q value directly. To allow using Q-learning algorithm in a more general state space, regardless of size and continuity, the DQN algorithm uses a constitutional neural network as a function approximation to estimate the Q function by $Q(s,a;\theta) \approx Q*(s,a)$, where $\theta$ is the network's weight parameters. At each iteration $i$, the DQN is trained to minimize the mean-square error (MSE) between the Q-network and $y=r+\gamma max_{a'}Q(s',a';\theta_{i}^{-})$, where $\theta_{i}^{-}$ is the network's weight from the previous iteration. The loss function In this approach can be expressed as follows: 
When in a domain with a state space that is fairly largeor continuous (e.g. Atari games or driving), it is not feasibleto directly compute the Q value. To allow the use of Q-learningalgorithm in a more general state space, regardlessof size and continuity, the DQN algorithm uses a constitutionalneural network as a function approximation to estimatethe Q function by $Q(s,a;\theta) \approx Q*(s,a)$ where $\theta$is the network's weight parameters. At each iteration, i, theDQN is trained to minimize the mean-square error (MSE)between the Q-network and $y=r+\gamma max_{a'}Q(s',a';\theta_{i}^{-})$ where $\theta_{i}^{-}$ is the network's weight from the previous iteration.The loss function in this approach can be expressed as
\begin{equation} \label{eq:loss}
L_{i}(\theta_{i})=E_{s,a,r,s'}[(y - Q(s,a;\theta_{i}))^{2}]
\end{equation}
where {$s, a, r, s′$} are state-action samples drawn from experience replay memory with a mini-batch of size 32. The reward $r$ is calculated using reward clipping that scales thescores by clipping all reward when positive at 1, negative at-1 and 0 when rewards are unchanged. The use of experiencereplay memory, a target network, and reward clippinghelps to stabilize the learning. To ensure the agent obtainssufficient exploration of the state space, DQN also uses anaction $\epsilon$-greedy policy. 
    
The usage of experience replay memory, a target network, and reward clipping help to stabilize the learning. To ensure the agent sufficiently explores the available state space, DQN also utilizes an action $\epsilon$-greedy policy.

\subsection{Pre-Training Networks for Deep Reinforcement Learning}
%Deep reinforcement learning generally needs to balance two tasks at the same time: (1) feature learning and (2) policy learning. Even though deep reinforcement learning has already been quite successful in learning both tasks in parallel, to ensure model converge and perform well it requires long training times and a large amount of data. To address feature learning task, we believe a supervised CNN model with human demonstration data input would dramatically speed up the learning process and quality, which from the leverage more resource on policy learning. Our work learns the feature representations by pre-training deep reinforcement learning's network using human demonstrations from non-experts. We refer to this approach as a \textit{pre-trained model}. \cite{Cruz_2017}
Deep RL generally needs to balance two tasks at the sametime: (1) feature learning and (2) policy learning. Eventhough Deep RL has already been quite successful at performingboth tasks in parallel, to ensure model convergenceand performance requires a long training time and a largeamount of data. To address the feature learning task, we believea supervised CNN model with human demonstrationdata input would dramatically speed up the learning processand quality, which from the leverage more resource on policylearning. In our work, deep RL learns feature representationsby pre-training its network using human demonstrationsfrom non-experts; we refer to this approach as a \textit{pretrainedmodel}. \cite{Cruz_2017}

%The pre-trained model method is similar to Bojarski's End-to-End approach\cite{Nvidia} with a deep CNN to learn the feature space. We also applied data augmentation to increase the sample size by adding artificial shifts and rotations. Unlike Bojarski's work, our work relies only on the center camera, and we also change the dimension of input, convolution filter size and network work output dimension to fit our approach. We construct the network as a multi-classification model. We assume here that humans provide correction action (labels) while driving. 
The pre-trained model method is similar to Bojarski'sEnd-to-End approach \cite{Nvidia}, using a deep CNN tolearn the feature space. We also applied data augmentationto increase the sample size by adding artificial shifts and rotations.Unlike Bojarski's work, our approach relies only thecenter camera, and we also changed the dimension of input,convolution filter size, and network work output dimensionto fit our approach. We construct the network as a multi-classificationmodel; we assumed that humans could providecorrection action (labels) while driving.

%The model parametrized by using: (1) MSE loss function, (2) Adam optimizer\cite{Kingma_2014} with a learning rate of 0.0001. The training library is Keras with Tensorflow backend. We use a batch size of 128. The CNN architecture follows the same structure with different hyper-parameters for input dimension, filter size, and output dimension. It includes a normalization layer, five convolution layer and each with dropout layer. It follows five flatten layer and each using a dropout layer. The activation function is ELU, the regulation function is L2. The original network's output has a single output for each valid action, which is not usable in our work. Instead, we increase the output dimension to three, which are throttle, steering and break. The learned weights and biases from the pre-trained CNN are used to initiate the DQN's network.
The model was parametrized by using an MSE loss functionand an Adam optimizer\cite{Kingma_2014} with alearning rate of 0.0001. The training library is Keras witha Tensorflow backend. We used a batch size of 128. TheCNN architecture followed the same structure, with differentparameters for its input dimension, filter size, and outputdimension. It included a normalization layer and fiveconvolution layers, each with a dropout layer. It followedfive flattened layers, each with a dropout layer. The activationfunction was ELU and the regulation function is L2.The original network's output had a single output for eachvalid action, which was not appropriate for our work. Instead,we increased the output dimension to three: throttle,steering, and brake. The weights and biases learned from thepre-trained CNN were used to initiate the DQN network.
    
%Since we are handling the raw image data, so the first layer of normalization is extremely important. Since normalization will help to generalize the model faster due to the different lighting captured from the camera. In all fully connected layers, we also applied normalization for the parameters that passed down the network. This normalization prevents learned parameter from either vanishing or exploding. The network has roughly 27 million trained weights and 250 parameters.
We were handling raw image data, so the first layer ofnormalization was extremely important, since normalizationhelps to generalize a model faster (due to the different lightingcaptured from the camera). We also applied normalizationfor the parameters that passed down the network in allfully connected layers. This normalization prevented learnedparameters from either vanishing or exploding. The networkhad roughly 27 million connections and 250 parameters.

\subsection{Interactive Deep Q-Network}
\label{sec:interactive-dqn}
As part of our work, we introduce the concept of human suggestion to the original DQN paradigm, which we call Interactive Deep Q-Network (IDQN). Recall that DQNs learn the optimal policy by first exploring the state space to learn rewards associated with visual input into the CNN used by the DQN. The visual input most often takes the form of a camera view, either of the state space or some small portion thereof. The Replay Memory stores tuples of these events, which take the form $[image, action, reward, done]$.

In order for an agent to discover the desired policy, the reward function must be properly set up such that the DQN can converge with discovered rewards. This means the reward policy can be tedious to build and must be reconfigured if some additional actions are desired. To solve this, we propose giving a human the ability to add suggestions to the agent in the form of adding extra tuples to the Replay Memory with elevated reward values for future training.

To accomplish this, a visual input system was designed to allow a trainer to suggest, either through a keyboard or GUI buttons, more appropriate actions the agent should take at a given point in time. When the trainer signals to add a suggestion, the last trained frame is re-added to the Replay Memory (but not the History, which is used for inference only) that will be later sampled from to continuously train the agent. 

Over time, the trainer's input is sampled against and due to its elevated reward shapes the policy the agent uses to incorporate the preferences the trainer is attempting to convey. The benefit is that the agent both learns the policy that maximizes the reward function at a faster rate as well as more complex policies that may only be known to the trainer providing the suggestions.

\begin{figure}
	\centering
	\includegraphics[width=\columnwidth]{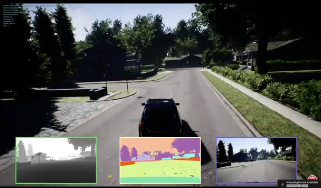}
    \caption{Microsoft AirSim simulator.}
    \label{fig:airsim}
\end{figure}
% ############################## Experiments ##############################
% Evaluate your system 
% If introduce algorithm -> rate, speed. Etc. 
% Identify: 
%  the evaluation metrics, baseline (why they are good baselines, because), what to compare with, what define success 
% success cases 
% experimental groups 
% Experimental methodology (how you test the experiment which is already working): how did you run, run experiment 10 times and average over 100 times, 80% training 20% test, etc 
% Should reflect the motivation and contribution from 1. 
\section{Experiment Design}
% Describe your approach in detail 
% Description of your idea, how it’s works 
% Two sessions: background information (2 paragraph RL)  
We use AirSim (\url{https://github.com/Microsoft/AirSim}), an open source simulator based on Unreal Engine as an autonomous vehicle agent \cite{shah2018airsim} \textbf{Figure \ref{fig:airsim}}. The deep reinforcement learning DQN and supervised learning CNN are both implemented using Tensorflow; the rest of the platform consisted of:

\begin{itemize}
	\item Windows 10 Pro x64
    \item AirSim Neighborhood Binary
	\item Python 3.6.3
   	\begin{itemize}
   		\item msgpack-rpc-python
        \item numpy
        \item opencv-python
        \item pillow
        \item tensorflow or tensorflow-gpu
   	\end{itemize}
	\item Microsoft Cognitive Toolkit
	\item Python Packages: (Install using pip3 install)
    \item CUDA 8.0 (GPU only) and cudNN 6 (GPU only)
\end{itemize}

\subsection{Supervised Convolution Neural Network}
%Due to limited computational resources and the time constraints of this project, we only used four datasets for the human demonstration. Yet we still achieved less than 0.1 and 0.3 for training and validation data respectively. The images we used are from the center scene image camera.
Due to limited computational resources and time constraints,we used only four datasets from human driving demonstrations.Regardless, we still achieved values $<$ 0.1 and 0.3for the training and validation data, respectively. The imagesused are from the center scene image camera.

\begin{figure}[h]
	\centering
	\includegraphics[scale=0.26]{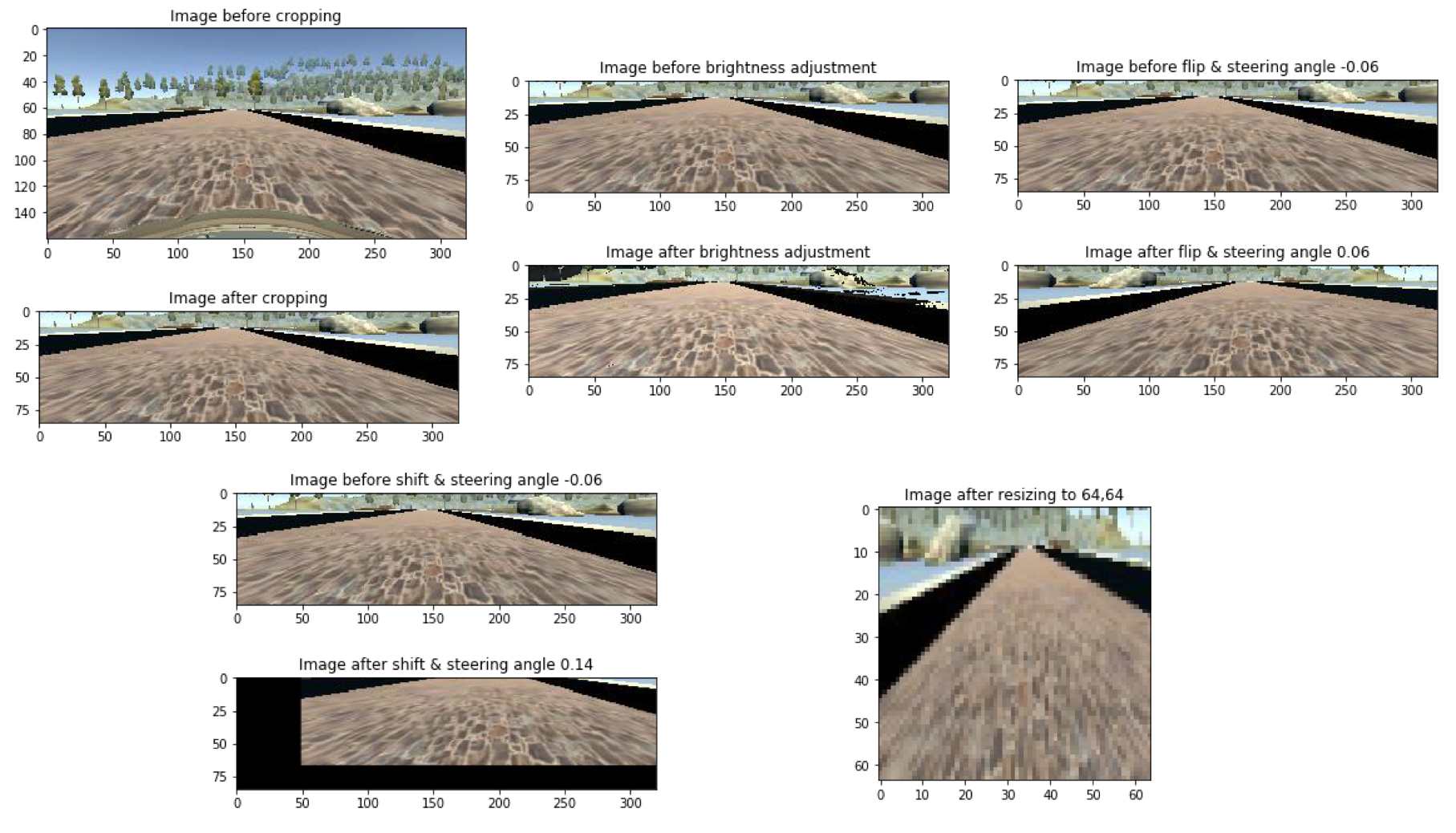}
    \caption{Image augmentation for CNN.}
    \label{fig:image-augmentation}
\end{figure}

%After collecting more than 1500 image frames from a human demonstration, we first constructed augmentation of the images, as shown in \textbf{Figure \ref{fig:image-augmentation}}. Since we assume the CNN model only focus on the lower part of the image, which is the road, the images were cropped accordingly. To mimic the real road condition, we also include artificial shifts and rotations to help the network to learn from poor position or orientation. The magnitude of these perturbations is randomly applied from a normal distribution with zero mean and the standard deviation is twice the standard deviation reported from the Bojarski's End-to-End approach paper.
After collecting more than 1500 image frames from humandemonstrations, we first constructed augmentation ofthe images \textbf{Figure \ref{fig:image-augmentation}}. Since we assumed the CNN modelwould only focus on the lower part of the image, the road,the images were cropped accordingly. To mimic real roadconditions, we also included artificial shifts and rotations tohelp the network to learn from poor position or orientationdata. The magnitude of these perturbations was randomlyapplied from a normal distribution with a mean of zero anda standard deviation twice the standard deviation reported inBojarski?s End-to-End approach paper.
    
%As mentioned earlier on, the CNN model has 5 convolution layer and four fully-connected layers (different from the Bojarski's work). After each of the layer, we applied a dropout layer to randomly remove a certain percentage of the learned parameters. For conventional layer, the dropout rates are all 0.5, and for fully connect layer the rates are 0.5, 0.4, 0.25 and 0. In this work, we also used an exponential linear unit as the activation function to include non-linearity, which is different from the original work. The batch size we have tested are 128 and 64, which didn't affect model performance much. 
As mentioned earlier, the CNN model has five convolutionlayers and four fully connected layers. We applied adropout layer after each layer to randomly remove a certainpercentage of the learned parameters. For conventional layers,the dropout rates were all 0.5, and for fully connectedlayer the rates were 0.5, 0.4, 0.25, and 0. In this work, wealso used an exponential linear unit as an activation functionto include non-linearity. The batch sizes tested were 128and 64; the batch size did not appear to have a large effecton model performance. Another difference from the originalpaper is that Instead of using three cameras--left, right, andcenter--we used only a center camera. because it's the onlyoption AirSim offers.
    \begin{table}
    \begin{center}
      \begin{tabular}{ | l | c | r }
    	\hline
        Layer & Dimension\\
        \hline
        Input & $64 \times 64 \times 3$ \\ \hline
        \hline
        Convolution2D & $24 \times 5 \times 5$ \\ \hline
        \hline
        Dropout & $0.5$ \\ \hline
        \hline
        Convolution2D & $36 \times 5 \times 5$ \\ \hline
        \hline
        Dropout & $0.5$ \\ \hline
        \hline
        Convolution2D & $48 \times 5 \times 5$ \\ \hline
        \hline
        Dropout & $0.5$ \\ \hline
        \hline
        Convolution2D & $64 \times 3 \times 3$ \\ \hline
        \hline
        Dropout & $0.5$ \\ \hline
        \hline
        Convolution2D & $64 \times 3 \times 3$ \\ \hline
        \hline
        Dropout & $0.5$ \\ \hline
        \hline
        Fully Connected & $1164$ \\ \hline
        \hline
        Dropout & $0.5$ \\ \hline
        \hline
        Fully Connected & $100$ \\ \hline
        \hline
        Dropout & $0.4$ \\ \hline
        \hline
        Fully Connected & $50$ \\ \hline
        \hline
        Dropout & $0.25$ \\ \hline
        \hline
        Fully Connected & $10$ \\ \hline
        \hline
        Fully Connected & $3$ \\ \hline
      \end{tabular}
      \caption{CNN Architecture}
    \end{center}
    \end{table}
    
\subsection{Deep Q-Network}

For the Deep Q-Network (DQN) portion of our experiment, the original network used in Minh's 2015 paper was used, sourced from an existing Python implementation included as an example in the AirSim GitHub repository. This code implemented the following DQN components:
\begin{itemize}
\item \textbf{Action Model} - CNN model used in action inference which is trained frequently (after 200 steps, then every 4 steps)
\item \textbf{Target Model} - CNN model used in loss calculations which is cloned from the Action Model occasionally (every 1000 steps)
\item \textbf{Replay Memory} - Holds up to 500,000 event tuples previously described which can provide mini-batches for training the Action Model
\item \textbf{History} - Holds N recent visual inputs for historical sequence inputs
\item \textbf{Linear Epsilon Annealing Explorer} - Scales the exploration rate of the agent based on a maximum random chance (100\%) that is phased down to a minimum random chance (5\%) over a given number of steps (5,000)
\item \textbf{DQN Agent} - Wrapper class that combines the other components with functions to pick an action based on the CNN policy approximator and exploration policy, save observations based on actions taken, and retrain the model based on sampled mini-batches from the Replay Memory.
\end{itemize}
The model is trained with a learning rate of 0.001, momentum of 0.95, and mini-batch size of 32 events. 

For this experiment, the input consisted of 84 x 84 images from the simulated front camera, converted to grayscale. The action space consisted of: forward (no turning), left (-0.25 steering), and right (0.25 steering) all at a constant acceleration of 0.35. The reward function, which can be seen in \textbf{Equation \ref{eq:reward}}, measures the distance from the center of the street, angle from the centerline following the street, speed traveling, whether the car has left the boundaries of the street, and whether a collision has occurred. 
\begin{equation} \label{eq:reward}
r(x) = \Delta d(x)+\angle c(x)+v(x)-e_{oob}-e_{collided}
\end{equation}
The last two measurements are considered catastrophic and result in an end to the episode and a large negative reward.

\subsection{Human Suggestion for Deep Q-Network}
\begin{figure}
	\centering
	\includegraphics[width=\columnwidth]{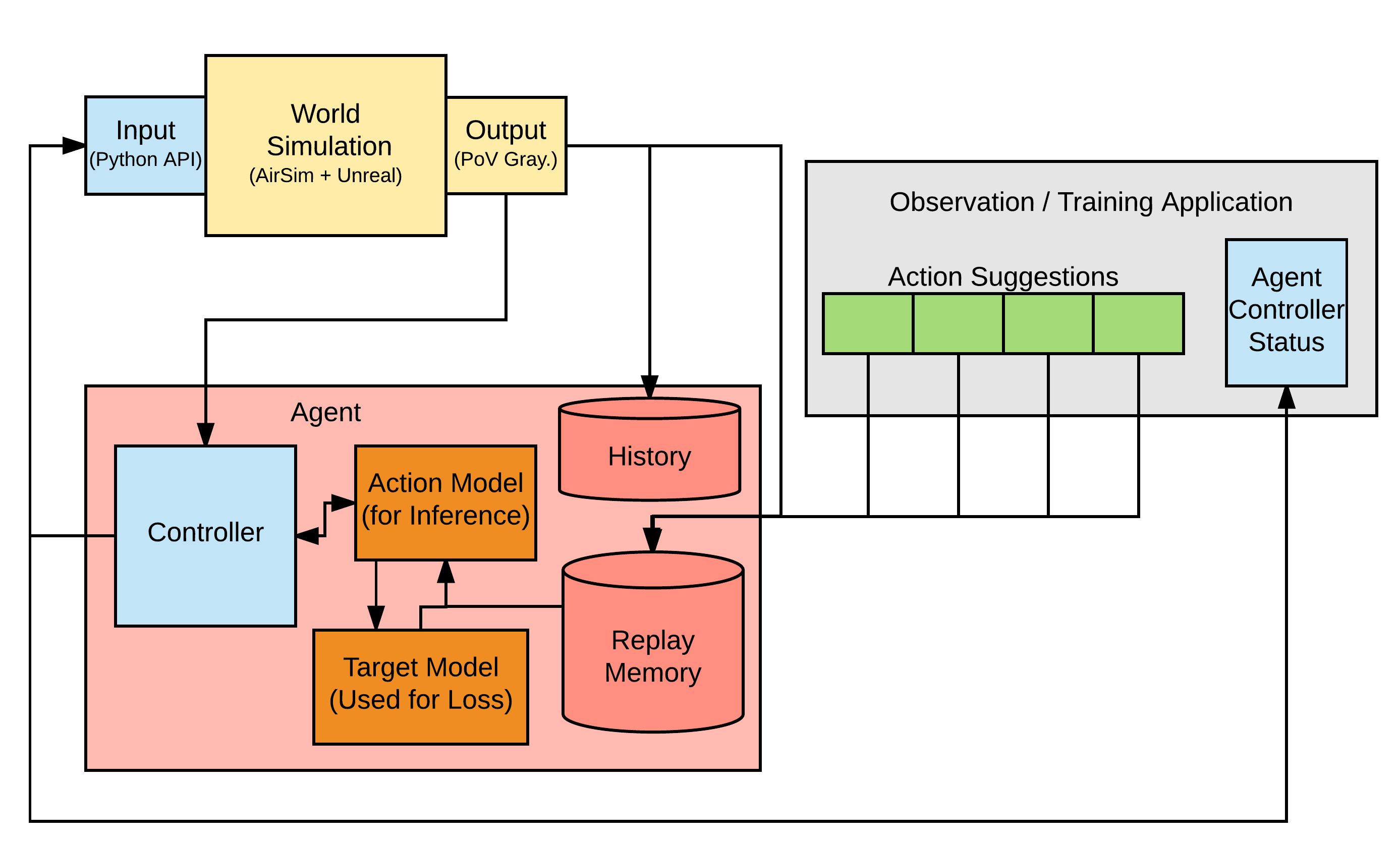}
    \caption{DQN Suggestion Pipeline}
    \label{fig:dqn-pipeline}
\end{figure}

In order to provide a mechanism for trainers to add suggestions during training, a GUI was developed which allowed the trainer to add suggestions to the agent during training. A representation of the pipeline which combines this GUI with the existing DQN agent and simulation system is shown in \textbf{Figure \ref{fig:dqn-pipeline}}. The training workflow follows the sequence:
\begin{itemize}
\item[1] Run AirSim simulator (choose Car mode)
\item[2] Run the app.py Python script
\item[3] Start the DQN Agent by pressing the space bar
\item[4] The agent continues to explore the simulation space, making actions and learning the policy
\item[5] (Suggestion Only) The trainer can use either the UP/LEFT/RIGHT arrow keys or the GUI buttons to suggest FORWARD, LEFT, or RIGHT actions respectively
\end{itemize}

\subsection{Evaluation Criteria}
%For model learning from human demonstration, the MSE loss was used as one criteria. As most supervised learning approach, the measurement of performance are a function of the difference between predict action and the human demonstrated action. Another measurement we included is the accident rate. The less accidents, the better learning is.
MSE loss was used as one criterion for model learning fromhuman demonstration. As with most supervised learningapproaches, the measurements of performance are a functionof the difference between the predicted action and thehuman-demonstrated action. Another measurement we includedis the accident rate: fewer accidents indicated betterlearning.

%For IDQN, we look at how the reward improves over the progression through episodes. Specifically, we measure the mean and standard deviation of the reward per episode, total reward gathered per episode, and the total number of steps taken per episode. We deep success as an improvement of the mean and standard deviation, which can indicate a more optimal policy found, as well as an increase in total reward per episode, which, if seen in concert with an increase in steps taken per episode, can indicate the episode lasted longer, meaning the car traveled further down the street.
For IDQN, we looked at how the reward improves overthe progression through episodes. Specifically, we measuredthe mean and standard deviation of the reward per episode,total reward gathered per episode, and the total number ofsteps taken per episode. We deem success as an improvementin the mean and standard deviation, which can indicatediscovery of a better policy, as well as an increase in total rewardper episode, which, if seen in concert with an increasein steps taken per episode, can indicate the episode lastedlonger, meaning the car successfully traveled further downthe street.

% ############################## Results + Discussion ##############################
% Should reflect the related work 
% Outline how everything do 
% Describe results 
% Discussion: draw natural conclusion from the results, aka, interpret the results 
% Whey it’s great, why you should care 
\section{Results}
%In the human demonstration part of the work, we achieved an average training loss of 0.1 and validation loss of 0.3. Which is quite an improvement considering only 4 datasets were collected. In the CNN, we used the exponential linear unit (ELU) as activation function in the CNN. The reason for doing so is that ELU not only help avoid a vanishing gradient via the identity for positive values, it also improve the learning  characteristic by included a negative values which allow it to push mean unit activation closer to zero. In essence, the ELU improved network and speed up training.
In the human demonstration part of the work, we achieved anaverage training loss of 0.1 and validation loss of 0.3, quitean improvement considering only 4 demonstration datasetswere collected. For the CNN, we used an exponential linearunit (ELU) as the activation function. This not only helpedavoid a vanishing gradient via the identity for positive values,it also improved the learning characteristic by includingnegative values, which allowed it to push the mean unit activationcloser to zero. In essence, the ELU improved thenetwork and sped up training. 

%One reason why we didn't achieve less than 0.001 loss as reported from the original Nvidia End-to-End learning paper is due to the extra predictions we included. Instead of single output, our CNN model return three results, throttle, break, and steering angle. We believe a better metrics could be loss divided by three, since the three output all contribute to the loss (presumably not equally). 
One reason we did not achieve a loss of less than 0.001--as reported in the original Nvidia End-to-End learning paper\cite{Nvidia}--is because of the extra predictions we included.Instead of a single output, our CNN model returnedthree results: throttle, brake, and steering angle. We believea better metric might be loss divided by three, since the threeoutputs all contributed to the loss (presumably not equally).
    
\begin{figure}[h]
	\centering
	\includegraphics[width=0.45\columnwidth]{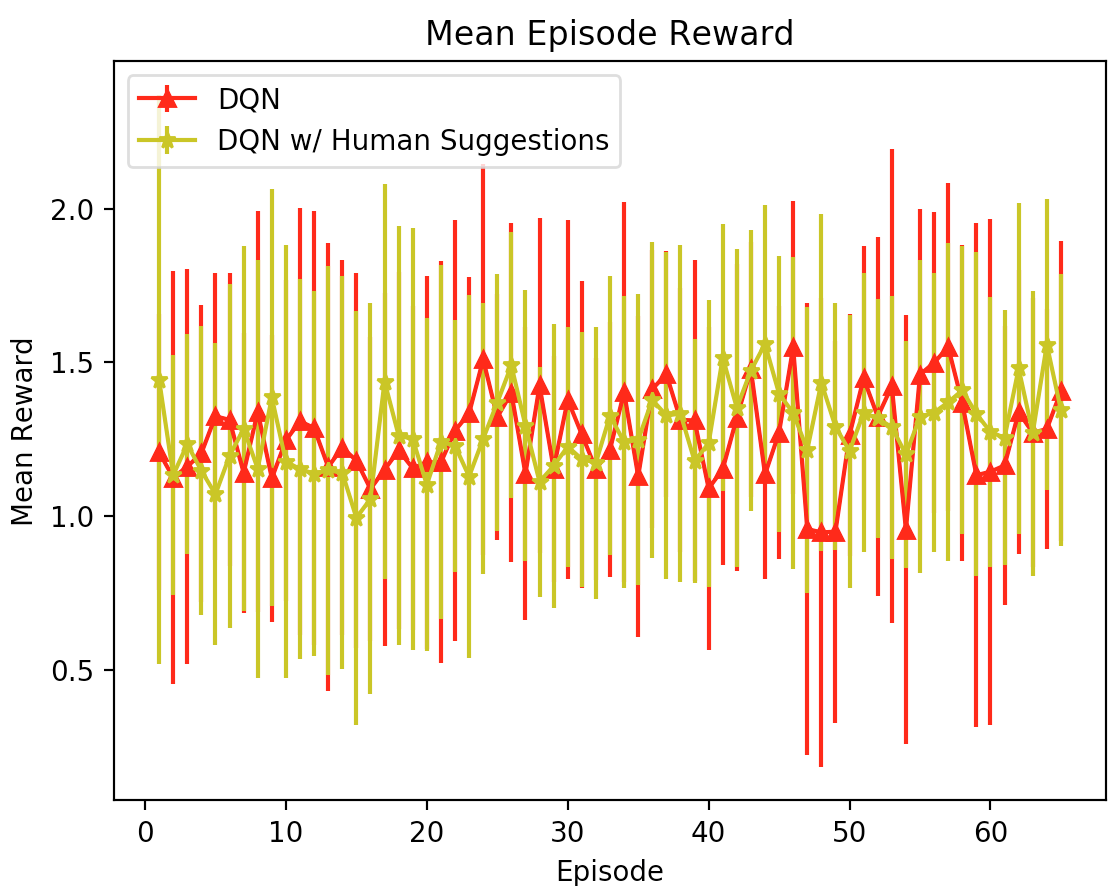}
	\includegraphics[width=0.45\columnwidth]{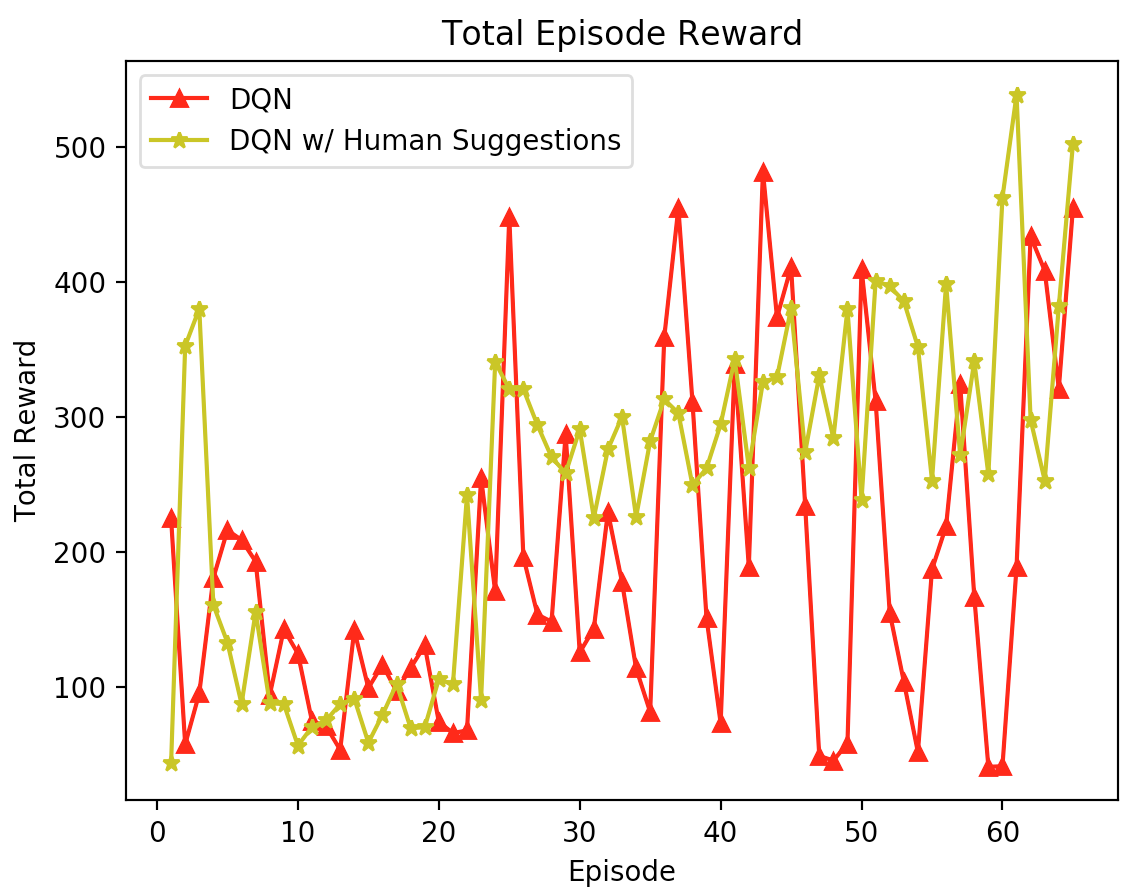}
	\includegraphics[width=0.45\columnwidth]{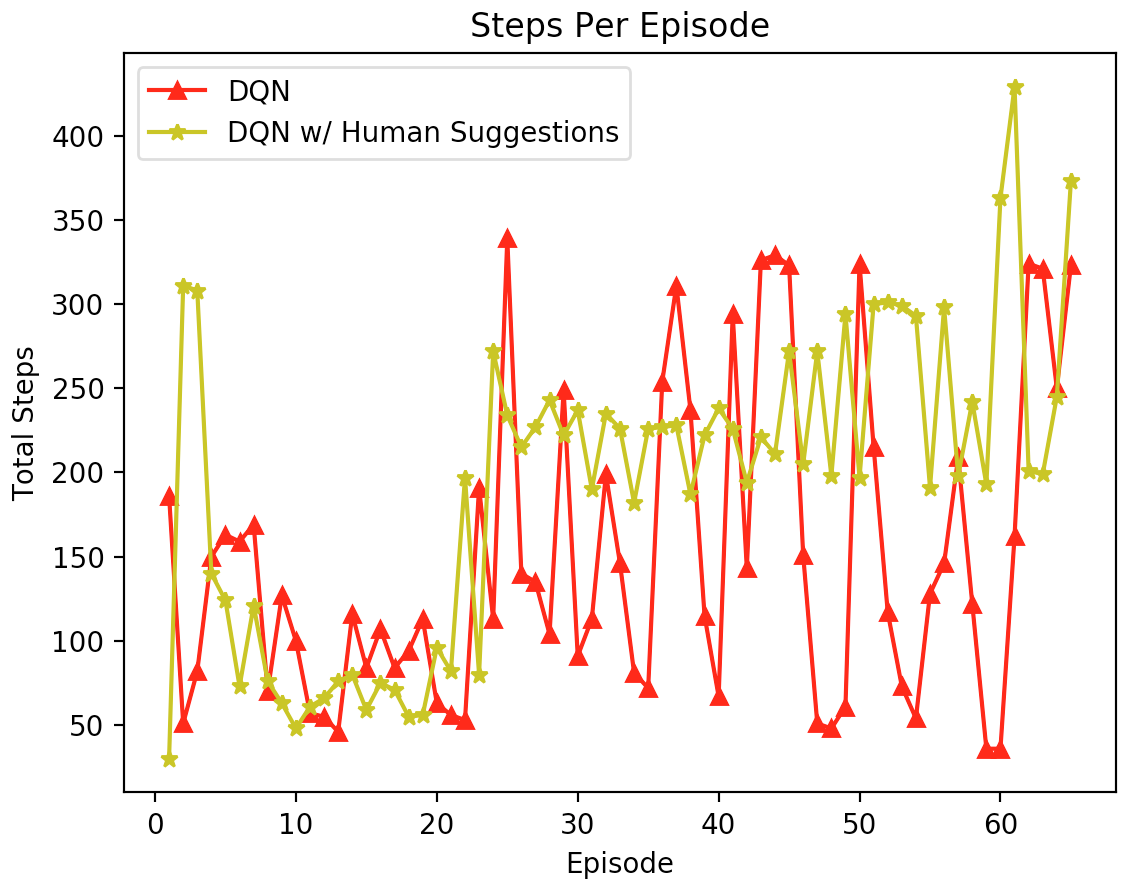}
    \caption{Results obtained from DQN vs. DQN with Human Suggestion (IDQN) showing Mean Episode Reward, Total Episode Reward, and Total Episode Steps Taken}
    \label{fig:dqn-results}
\end{figure}

Results from the IDQN experiment, shown in \textbf{Figure \ref{fig:dqn-results}}, show some improvement in the form of a tightening of the standard deviation in mean reward per episode. We take this to mean that the policy has converged on a more optimal approximation of the reward policy. In addition, we see that the total reward and total steps have both seen a measured increase, which, as noted previously, we take to mean the agent can travel further in the episode and thus gather more reward.

More generally, we saw the ability for the agent to learn more complex policy approximations. This was shown by the agent learning, though suggestions given to the agent during train, learning to make a left-hand turn at an intersection. In comparison, the original DQN agent failed to learn what navigation to perform at the intersection and simply ran into the fence at the opposite side of the street. This resulted in much larger total rewards/steps seen in three of the last four episodes in the IDQN agent in the results in \textbf{Figure \ref{fig:dqn-results}}.

\section{Conclusion}

%We weren't able to include the pre-trained mode in the Deep Q-Network, due to the complexity of the simulator environment and time strain. However, from the results of individual perform of the two models, we believe our approach would be feasible.
We were not able to include the pre-trained mode in theDeep Q-Network, due to the complexity of the simulatorenvironment and time strain. However, from the results forindividual performance of the two models, we believe ourapproach is feasible.

%Human demonstrations are a success our approach (partially). It is important to understand how the demonstrator's performance and the amount of demonstration data affect the benefits of pre-training the network in future work. Also we could use human demonstration End-to-End learning as a comparison candidate for our approach. Although the suggestion of demonstration hour is more than 100 driving hours, but from our work, we believe a much smaller sample size could achieve a similar results. In the pre-trained CNN model, we ignore the information collected from the depth and segmentation cameras. There are several studies that show that these information could further improve self-driving agent.
Human demonstrations are partially responsible for thesuccess of our approach. It will be important to investigatehow the demonstrator's performance and the amountof demonstration data affect the benefits of pre-training thenetwork in future work. Human demonstration End-to-Endlearning could also be used as a comparison candidate forour approach. Although the suggestion for demonstrationhours is more than 100 driving hours, from our work, webelieve a much smaller sample could achieve similar results.In the pre-trained CNN model, we ignore the informationcollected from the depth and segmentation cameras. Severalstudies have shown that this information could further improveself-driving agents.

We also show that our IDQN approach is successful in increasing the speed and accuracy of training over the original DQN implementation. Additionally, the IDQN approach allows for the learning of more complex policy approximation to be learned without rebuilding a more complicated reward function to instruct the agent. In the future, we hope to take some direction from the work by Hausknecht \cite{hausknecht2015deep} involving Deep Recurrent Q-Learning, which involves adding an LSTM layer to extract knowledge from sequential images used for input. This has the added benefit of learning policy in partially observable MDPs like the forward-facing camera used for input in the experiments conducted as part of our work. We also plan to investigate methods to limit the ability for the trainer to "over-train" by providing too many suggestions without seeing the appropriate feedback in the form of better results. This could be from either better instructions or better order of training from user suggestions for better transparency.

\bibliographystyle{IEEEtran}
\bibliography{sources}

\end{document}